\documentclass[11pt]{article}

\usepackage[T1]{fontenc}
\usepackage[utf8]{inputenc}
\usepackage{graphicx}
\usepackage{caption}
\usepackage{subcaption}
\usepackage{tabularx}
\usepackage{algorithm,algpseudocode}
\usepackage{url}
\usepackage{color}
\usepackage{times}
\usepackage{booktabs}
\usepackage{amsmath}

\definecolor{rulecolour}{rgb}{0.0,0,0.0}
\definecolor{headingcolour}{rgb}{0,0,0.0}

\newcommand{\assign}{\ensuremath{\leftarrow}}

\graphicspath{{./pix/},{../pix/}}

\usepackage{amssymb}
\usepackage{authblk} % For author and affiliation management
\usepackage{fancyhdr}

\fancypagestyle{firstpage}{%
    \fancyhf{}%
    \fancyfoot[C]{\small This work has been submitted to the IEEE for possible publication. Copyright may be transferred without notice, after which this version may no longer be accessible}%
}
\begin{document}   

%\begin{frontmatter}



\title{Actionable and diverse counterfactual explanations incorporating domain knowledge and plausibility constraints}


\author[1]{Szymon Bobek \texttt{szymon.bobek@uj.edu.pl}}
\author[2,3]{Łukasz Bałec \texttt{l.balec@cyberfolks.pl}}
\author[1]{Grzegorz J. Nalepa \texttt{grzegorz.j.nalepa@uj.edu.pl}}

\affil[1]{Faculty of Physics, Astronomy and Applied Computer Science, Institute of Applied Computer Science, Jagiellonian Human-Centered AI Lab}
\affil[2]{cyber\_Folks\textsuperscript{\texttrademark}, Krakow, Poland}
\affil[3]{FreshMail Sp. z o.o., Krakow, Poland}
       
            \maketitle
            \thispagestyle{firstpage}

\begin{abstract}
Counterfactual explanations improve the actionable interpretability of machine learning models by identifying minimal changes required to achieve a desired outcome. However, existing methods often neglect dependencies among features, which can lead to unrealistic or impractical modifications. This limitation reduces the usefulness of counterfactual explanations in real-world decision-support systems.

Motivated by applications in cybersecurity for email marketing, we propose DANCE (Diverse, Actionable, and Knowledge-Constrained Explanations), a method for generating counterfactuals that incorporate feature dependencies and domain constraints. DANCE models relationships between features using linear and probabilistic structures that can be learned from data or specified by experts. These dependencies are enforced during the search process to improve plausibility and feasibility.

The method jointly optimizes plausibility, diversity, proximity, and sparsity within a unified objective. We evaluate DANCE on 140 datasets from OpenML and demonstrate that it achieves competitive or superior performance compared to existing approaches across multiple evaluation criteria. Additionally, we validate the method in a real-world industrial setting in collaboration with an email marketing platform, showing that it produces domain-consistent and actionable recommendations.

\end{abstract}

\section{Introduction}
Explainable Artificial Intelligence (XAI) has gained substantial momentum in recent years, attracting attention from both researchers and various business sectors, including healthcare, heavy industry, and legal firms.
%A multitude of XAI methods were developed, each addressing different facets of explaining AI system decisions.
Among various explanation paradigms, counterfactual (CF) and contrastive explanations are often the most valued by stakeholders such as end-users and domain experts~\cite{buts2022understandability,miller2019social,mushroom} because they give explicit examples of which features play the most influential role in AI models' decisions and how their values should be changed in order to modify the overall AI system output.
However, CF explanations that conflict with established domain knowledge or common sense are particularly problematic, as they fail to provide actionable insights and reduce trust in the AI system.
For example, a CF explanation suggesting that reducing the risk of diabetes involves altering parameters such as sex, age, and pregnancy status would be deemed impractical and irrelevant, given the well-established medical understanding of diabetes management and common sense.
Despite this, only a minority of existing methods for generating CFs consider plausibility explicitly and in a way that is auditable and allow human modification.
This issue has become one of the emerging challenges in the XAI area, not only limited to counterfactual explanations~\cite{causality}, but most often in practical implementation of XAI methods to different business cases.

Our work is on the one hand directly rooted in the very active area of basic CF research in XAI (cf. Section~\ref{sec:sota}), yet on the other hand is motivated by practical challenges in a specific domain, which is cybersecurity in online marketing (cf. Section~\ref{sec:motivation})\footnote{This work was supported in part by the Project "SendGuard—Improving the Security of Recipients Innovative Tool Based on Machine Learning Technology and Artificial Intelligence to Fight with the Problem of Spam and Phishing in Marketing and Transactional E-Mail Messages" co-financed from the Funds of European Regional Development Fund under Project POIR.01.01.01-00-0202/19-02; The research has been supported by a grant from the Priority Research Area (DigiWorld) under the Strategic Programme Excellence Initiative at Jagiellonian University.}.
In this case the ability to generate domain-informed, plausible counterfactuals stands to substantially enhance the relevance and quality of email campaigns, newsletters and communication with customers.
A counterfactual should provide actionable insights, suggesting specific alterations to a campaign (e.g., changing the subject line length, adjusting the send time, or refining the image-to-text ratio) to improve its perceived quality. 
However, the recommended modifications that do not align with industry best practices, contradict known audience preferences, or violate brand guidelines are unlikely to be adopted by practitioners, thus limiting their practical utility.
%Our work stems from a collaboration with Freshmail~\footnote{http://freshmail.pl}, the largest email marketing company in Poland, and their real-life business case that gave motivation for our research is described in Section~\ref{sec:motivation}.

In this work, we address this challenge by introducing DANCE (Diverse, Actionable, and Knowledge-Constrained Explanations), a method that explicitly models and enforces dependencies between features during counterfactual generation (cf. Section~\ref{sec:method}). These dependencies can be learned from data using directed graphical models or specified by domain experts. Importantly, we treat these structures as plausibility constraints rather than universally valid causal models, allowing the method to remain applicable across heterogeneous datasets.

DANCE integrates these constraints into a multi-objective optimization framework that simultaneously accounts for correctness, proximity, sparsity, diversity, and plausibility. The optimization is performed using a Tree-Structured Parzen Estimator, enabling efficient exploration of complex search spaces with conditional dependencies.

We evaluate DANCE in two settings. First, we conduct a large-scale benchmark on 140 datasets from OpenML~\cite{OpenML2025} to assess general performance across diverse domains (cf. Section~\ref{sec:eval}). Second, we present a real-world case study in email marketing, where plausible counterfactuals are essential for producing actionable recommendations aligned with domain practices (cf. Section~\ref{sec:userstudy}).

The results demonstrate that incorporating structured feature dependencies leads to counterfactual explanations that are more realistic, interpretable, and useful in practice.

% The rest of the paper is composed in the following manner.
% In Section~\ref{sec:sota}, we present the state-of-the-art solutions in the field of counterfactual generation and highlight the research gap that our work aims to address.
% Section~\ref{sec:motivation} outlines the real-life business case motivation for our work.
% The method we propose is introduced and described in Section~\ref{sec:method}.
% The evaluation of the method, including both quantitative and qualitative analyses on a real dataset, is presented in Section~\ref{sec:eval} and Section~\ref{sec:userstudy}, respectively.
% Finally, the summary of our findings and potential future research directions are discussed in Section~\ref{sec:summary}.

%Let us now move to the discussion of the state-of-the-art solutions in the field of counterfactual generation and highlight the research gap that our work aims to address.

\section{Related works}
\label{sec:sota}
%overview of CF generation methods
The XAI research area has experienced a blooming period over the recent years, which was driven both by the researchers who noticed the urgency of providing more control over the AI systems operation~\cite{rusellcontroll}, as well as legal authorities through the AI Act in EU~\cite{aiact}, and DARPA challenge~\cite{darpa} in the USA.
The full spectrum of XAI methods is vast and has been extensively covered by several surveys~\cite{guidotti2022cf,xaisurv,adversarial}, to which we refer readers for a comprehensive overview of the field. 
In this work, we focus specifically on counterfactual explanations, a distinct subfield within XAI that provides example-based explanations~\cite{examplebased}.

In this type of XAI method, the explanations are presented to the user as examples of specific instances, which may be real or artificially generated. 
For tabular data, the explanation takes the form of a row of feature values; in the case of image data, it is an image; and for text data, it is a text snippet, etc. 
Counterfactual explanations further ensure that the instance belongs to a contrasting class (potentially defined by the user) relative to the instance being explained.

Other methods for example-based explanations include adversarial examples~\cite{adversarial}, prototypes and critiques~\cite{mmd}, and explanations based on prototypical parts~\cite{protopnet}, among others. 
What distinguishes counterfactual explanations from these types is that counterfactuals are expected to meet certain additional properties. 
The most common properties are correctness, diversity, sparsity, proximity, and plausibility~\cite{guidotti2022cf}. 
However, no existing methods optimize for all these properties simultaneously; instead, they typically focus on just a subset.

Furthermore, very few methods prioritize the plausibility of counterfactuals. 
Those that do often use simple range-based constraints, which limit the ability to capture more complex relational constraints - this is precisely where our work focuses. 
Additionally, these methods typically do not permit expert interaction with the constraints, thereby disabling domain-knowledge-based constraints and relying solely on density-based and data-driven constraints to model plausibility.
In the following paragraphs, we provide a more in-depth analysis of existing methods with respect to correctness, diversity, sparsity, proximity, and plausibility criteria.
Our focus is on methods that address plausibility in some form or are considered significant milestones in the counterfactual generation landscape. 
A comparative summary of these methods is provided in Table~\ref{tab:cf_methods_comparison}. 
For a comprehensive review of all methods in the field, we refer the reader to~\cite{guidotti2022cf}.

One of the earliest counterfactual generation methods was Wachter~\cite{wachter}. 
It used a simple approach that minimized a loss function composed of two terms: one for the proximity of the counterfactual and the other for its correctness. This basic loss function was optimized using the gradient-based algorithm. 
The Wachter algorithm is often regarded as a baseline for other methods, as it was both the first and one of the most straightforward approaches to counterfactual generation.

DCE~\cite{dce} followed a similar idea as introduced in WACH, but optimizing a custom cost function that accounted for sparsity and proximity, but added component responsible for keeping the counterfactual coherent.
The authors defined coherence as consistency with the underlying data structure, which was limited to ensuring that the counterfactual generation method does not set all one-hot-encoded features to either \emph{0} or \emph{1}.
In this sense, we can consider this approach as a minimal plausibility constraint.
DCE was designed only to work with linear models.

One of the most well-known methods in the area of CF is called Diverse Counterfactual Explanations (DICE)~\cite{dice}. 
It further extends the ideas of Wachter and DCE by including additional terms in the loss function to ensure the broader spectrum of CF properties.
It was designed to handle diversity, sparsity, and proximity of counterfactual with a custom loss function that integrated all of the three criteria in a form of losses and optimized with different forms of search algorithms such as genetic algorithms, random search, gradient optimization. 
It utilized hinge loss to decide on the correctness of the counterfactual.
DICE does not directly account for plausibility in the objective function but allows for predefined ranges of feature values, which can help prevent implausible counterfactuals, such as values outside the possible scale or basic monotonicity constraints (e.g. age cannot decrease).
A similar approach to handling implausible feature values was proposed in MACE~\cite{mace} and CFNOW~\cite{cfnow}.
None of the aforementioned methods consider complex cases, where the constraints are conditional, or in some kind of relation (e.g. linear) with respect to each other.

Another group of methods approaches plausibility through representation learning. 
This includes algorithms such as CEM~\cite{cem}, EBCF~\cite{ebcf}, and CEGP~\cite{cegp}, or FRACE~\cite{frace}. 
These methods rely on autoencoders to ensure that the reconstruction loss of a newly generated sample (counterfactual candidate) is minimal, confirming that it is plausible—that is, that it belongs to the representation learned by the autoencoder during training.

Finally, the last group of methods ensures the plausibility of newly generated samples by positioning them in areas of high-density points or aligning them with the underlying data distribution.
This group can be further divided into endogenous explainers, which provide explanations by selecting existing samples from the given dataset (such as LUX~\cite{lux}, LORE~\cite{lore}), and exogenous explainers, which generate artificial samples such as DACE~\cite{dace}, OCEAN~\cite{ocean}, or PPCF~\cite{ppcf}.

% DCE~\cite{dce} CEM~\cite{cem}, EBCF~\cite{ebcf}, DACE~\cite{dace}, MACE~\cite{mace},  OCEAN~\cite{ocean}, CEGP/CFProto~\cite{cegp}, LUX~\cite{lux}, LORE~\cite{lore}, , Wachter~\cite{wachter}, cfnow~\cite{cfnow}

\begin{table*}[ht]
    \centering
    \caption{Comparison of representative counterfactual explanation methods. Columns indicate whether a method explicitly optimizes a given property (\textit{Yes}), supports it indirectly (e.g., by repeated runs), or does not address it. \textit{Plausibility} summarizes the main mechanism used to encourage feasibility (bounds/constraints, density, representation learning, or structured plausibility constraints).}
    \label{tab:cf_methods_comparison}
    \fontsize{7}{9.6}\selectfont
    \begin{tabularx}{\textwidth}{p{1.2cm}p{2cm}p{0.9cm}p{0.9cm}p{0.9cm}Xp{1.5cm}X}
        \hline
        \textbf{Method} & \textbf{Optimization Mechanism} & \textbf{Diversity} & \textbf{Sparsity} & \textbf{Proximity} & \textbf{Plausibility} & \textbf{Model-Agnosticism} & \textbf{Expert-Based Constraints} \\
        \hline\hline
        DICE~\cite{dice} & Various& Yes & Yes & Yes & Constraint-based & Yes & Range-based \\
        \hline
        DCE~\cite{dce} & Linear programming & Yes & Yes & Yes & Minimal & Linear models only & No \\
        \hline
        CEM~\cite{cem} & FISTA optimizer & No & No & Yes & Autoencoder-based & Yes & Range-based \\
        \hline
        EBCF~\cite{ebcf} & Gradient-based & No & No & Yes & Autoencoder + manually encoded plausibility constraints & Yes & Yes \\
        \hline
        DACE~\cite{dace} & CPLEX optimizer & No & No & Yes & Density-based / outlier-based & Linear and tree-based & No \\
        \hline
        MACE~\cite{mace} & SAT solvers & Yes & Yes & Yes & Constraint-based & No & Range-based \\
        \hline
        OCEAN~\cite{ocean} & Custom loss function & Not natively & No & Yes & Density-based / outlier-based & Only for tree ensembles & Logical and Linear \\
        \hline
        CEGP~\cite{cegp} & FISTA + K-D Tree search & No & Yes & Yes & Autoencoder-based & Yes & No \\
        \hline
        PPCF~\cite{ppcf} & Gradient-based & No & No & Yes & Density-based & Requires gradients & No \\
        \hline
        LUX~\cite{lux} & Entropy and distance & Not natively & No & Yes & Density-based & Yes & No \\
        \hline
        LORE~\cite{lore} & Entropy and distance & No & No & Yes & Density-based & Yes & No \\
        \hline
        Wachter~\cite{wachter} & Gradient-based & No & Yes & Yes & No & Yes & No \\
        \hline
        CFNOW~\cite{cfnow} & Tabu search & No & Yes & Yes & No & Yes & No \\
        \hline
    \end{tabularx}

\end{table*}

The motivation for our work, which stemmed from the visible gap in the landscape of state-of-the-art counterfactual explainers, was additionally backed by the practical needs.
A detailed description of the business motivation is given in the following section.

\section{Motivation grounded in practical use case from cybersecurity in email marketing}
\label{sec:motivation}

In the rapidly evolving digital landscape,  Communication Platform as a Service (CPaaS) providers have gained critical importance by enabling businesses to efficiently engage with their audiences through various communication channels. 
Among these channels, email marketing remains a fundamental tool for customer outreach, brand awareness, and relationship building. 
Modern CPaaS solutions offer comprehensive platforms for creating, managing, and distributing large-scale email campaigns. 
However, ensuring that these messages are both welcomed by recipients and meet stringent quality standards poses a significant challenge. 
The delicate balance between delivering valuable, personalized promotions and avoiding unwanted or irrelevant content~-- colloquially known as spam --~lies at the core of the email marketing domain.

Furthermore, there is a growing trend in cybersecurity, to fight misinformation in all possible channels.
Thus, carefully engineering spam messages that could be injected into a viable marketing campaign can cause an important security threat.
Although in the work described in this paper we consider content quality on a general level, and do not address cybersecurity directly, the technology we developed within the \emph{Sendguard} project with Freshmail is in fact security-oriented~\cite{bazan2025frm}.
The broader R\&D effort motivating this work aimed at developing AI-based techniques to strengthen the security of company mail servers by protecting them from sending spam or phishing messages. The project was co-financed by the Polish National Centre for Research and Development (NCBiR) between 2020-2023

Freshmail offers its customers a diverse platform for email marketing.
It is most commonly used with so-called marketing campaigns that their customers design with their marketers.
A campaign is first designed around a given topic, with a number of factors regarding both content and its form identified and configured.
Furthermore, possible target groups of campaign recipients are identified.
Using such a configuration, and additional schedules based on expert knowledge (e.g. when specific target groups tend to read their emails), large batches (tens or even hundreds of thousands) of email messages are sent.
Quality of messages is important not only from marketing point of view, but also security.
Should there be an increased level of low quality messages sent, the company servers could be put on blacklists by backbone Internet providers and effectively blocked from operation.

In this work, to quantify undesirability of email messages, we adopted an engagement-centric metric: an extremely low email open rate (e.g., less than 2\%) suggests that recipients do not find value in the content. While straightforward in principle, this approach surfaced deeper issues. 
The open rate alone could not fully explain the complex interactions among countless features that influence campaign performance. 
Factors such as subject line specificity, visual composition, send time, and audience segmentation all interweave to determine how a campaign is received.

This multifaceted nature of campaign data, enriched by detailed server logs, event records, and behavioral analytics, made classification and prediction profoundly challenging. 
Consequently, understanding why a model identified a campaign as undesirable was nearly as difficult as classifying the campaign itself.
This realization underscored the necessity for the use of XAI methods. 
Introducing XAI techniques enabled us to decipher the reasoning behind model predictions, illuminating which features or interactions influenced the classification of a campaign as spam-like. 
Such a transparency is instrumental in guiding both model refinement and practical improvements to the campaigns themselves.
Instead of simply labeling a campaign as undesirable, explanations, such as counterfactuals, guide marketers in transforming an ineffective campaign into one that engages and resonates with recipients.

While counterfactual explanations hold significant promise, existing generation methods are often limited by restrictive assumptions.
Standard approaches frequently overlook the complex joint distributions and dependencies present in real-world datasets. 
In real world scenarios, unrealistic or impractical changes proposed by counterfactuals limit their adoption and business value.
From the company perspective,
the ability to generate domain-informed, plausible consistent counterfactuals stands to substantially enhance the relevance and quality of email marketing campaigns. 
Having this requirement as a practically grounded motivation, our research highlights a path forward for integrating complex data-driven models with domain expertise to produce results that are not only correct in theory but also valuable in practice.
More broadly, this work fits the growing intersection of explainable AI and networked online systems, where model outputs translate into interventions that affect how information is delivered, perceived, and acted upon at scale. In such settings (including email marketing and adjacent platform pipelines such as recommendation, targeting, and moderation), explanations must be not only diverse and close to the original instance, but also feasible under structured dependencies among user, content, and context variables. Our dependency-graph constraints can be viewed as a lightweight network-structured prior over features, enabling counterfactual recourse that remains aligned with operational rules and observed conditional regularities. This perspective connects counterfactual explanation research with practical questions of trustworthy decision support in social systems, where transparency and actionability are prerequisites for adoption and responsible use.
This complements ongoing efforts to make AI systems deployed in large-scale digital communication pipelines more transparent and practically trustworthy.

In the following sections, we detail the algorithmic foundations of our approach, followed by empirical evaluations validating its benefits, and the broader implications for the future of AI-driven marketing communications and its security.

\section{Method description}
\label{sec:method}
%excact description of the method -- algorithsm, stages, theory etc: a) 
%Show how to model/discover the knoweldge,
%Show how to use it
%Show that we are better

\subsection{Background}
\label{sec:bkg}

DANCE allows us to learn the structure and magnitude of dependencies between features used by the blackbox model in the form of linear or nonlinear constraints on the feature values.
This structure is then used to generate counterfactuals consistent with it, and therefore more plausible and actionable.
Alternatively, the structure and/or specific constraint values can be provided in advance by an expert if they are known \textit{a priori}.

\begin{figure}[htb]
\centering
\includegraphics[width=1\columnwidth]{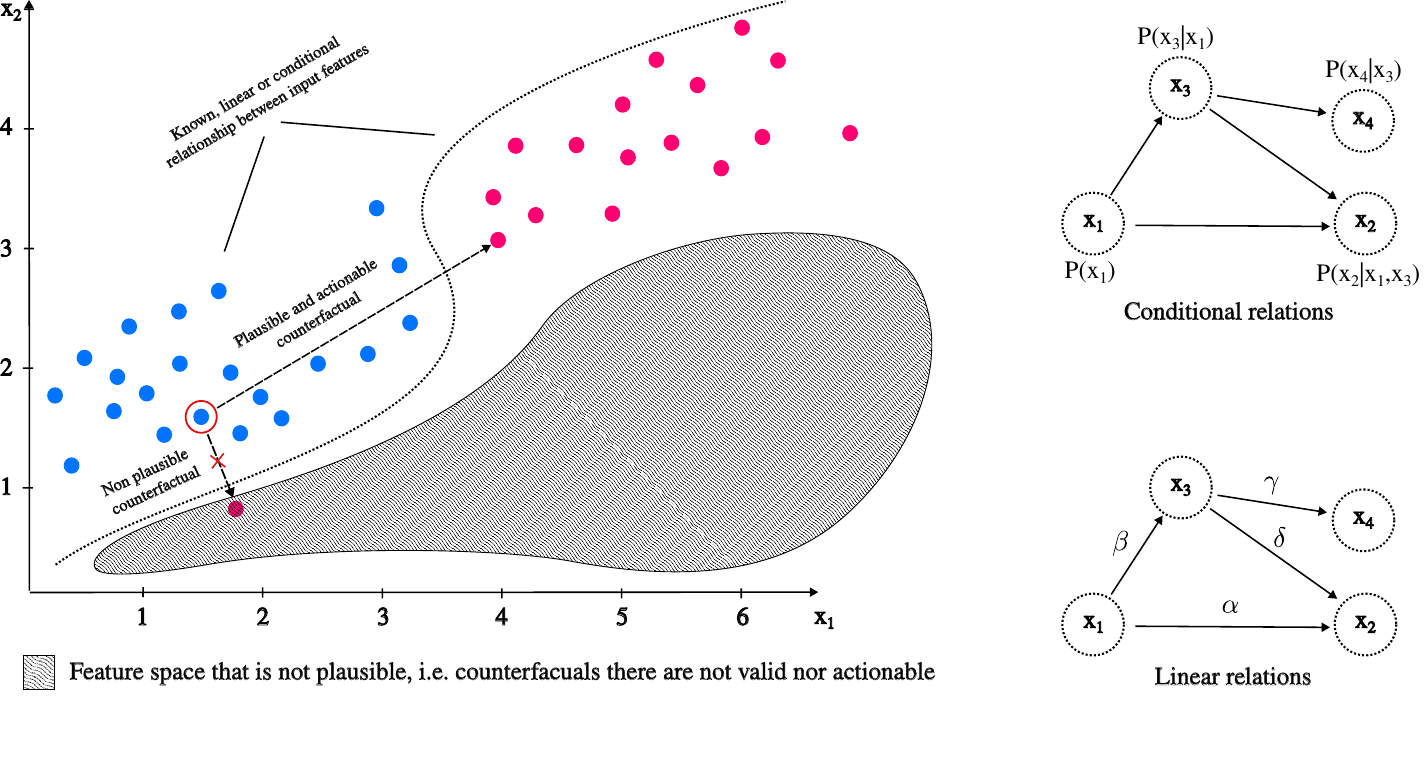}
\caption{Illustration of inter-feature plausibility constraints targeted by DANCE. A counterfactual that changes one variable in isolation may become infeasible when other variables are structurally linked (linear dependencies), and conditional constraints can exclude regions of the feature space (e.g., forbidden combinations). DANCE incorporates such constraints via a directed relationship graph and penalizes violations during search.}
\label{fig:case}
\end{figure}

We are focusing on two types of relation between features: a linear relation and possibly nonlinear conditional relations.
As the CF by definition is an instance that is produced by altering features values of the original instance, the relations ensure that modifications are consistent with given relations.
We employ a custom cost function to ensure not only plausibility, but also diversity and sparsity of explanations given in equation below and employed Tree-structured Parzen Estimator optimization algorithm to minimize it~\cite{parzen}. 
Figure~\ref{fig:case} shows the cases that our method handles. 

The linear relation between features prevent the method from altering only one feature value without changing the value of another one. 
The relation can be provided in a form of directed acyclic graph, where nodes represent variables and edges contain information about the linear relationship between them. 
This can be used to describe linear relationships such as \textit{"Increasing the ambient pressure, increases the temperature of boiling of a liquid".}

Conditional dependencies allow one to model more complex constraints that are non-linear, e.g. mark "forbidden" feature space areas.
Similarly to the previous type of constraints, this type is also defined by the directed acyclic graph, however, the nodes contain conditional probability distribution of feature value given the feature values of nodes from incoming edges. 
This in fact can be modeled in the form of a Bayesian network and can handle relationships that are more complex than simple linear constraints such as the aforementioned forbidden areas in the feature space: \textit{"Given the person is a man, the probability of him being pregnant equals 0"}.

%%%%%%%%%%%%%%%%%%%%
%TODO: details on how the graph is created.

%%%%%%%%%%%%%%%%%%%%%

The core of our approach was to design and optimize a loss function in a way that will evenly value different properties of final set of counterfactuals such as fidelity, proximity, sparsity, plausibility, and diversity.
For the design of the loss function, we build on the concepts introduced by Wachter and DICE, integrating different properties as separate components of the function, as shown in Equation~(\ref{eq:ccfloss}). 

\begin{equation}
\label{eq:ccfloss}
\begin{aligned}
L \;=\;& \overbrace{\max \left[ 0, \max_{c \neq d} ( \hat{P}_{i,c} - \hat{P}_{i,d}) \right]}^{
\parbox{.46\linewidth}{\centering\tiny Maximum of zero and difference between maximum probability of desired class $d$ versus other classes.}}
\\
&\;+\; \underbrace{\lambda \cdot \left( \omega_{\text{d}} L_{\text{div}} + \omega_{\text{x}} L_{\text{prox}} + \omega_{\text{s}} L_{\text{spars}} + \omega_{\text{p}} L_{\text{plaus}} \right)}_{
\parbox{.50\linewidth}{\centering\tiny Diversity, Proximity, Sparsity and Plausibility losses with weights. $\lambda$ is a global weight.}}
\end{aligned}
\end{equation}

%Here, $x$ is the instance being explained, $d$ is the desired class, and $\hat p_c(\cdot)$ denotes the black-box probability for class $c$. We minimize $L$; therefore each component is defined as a non-negative penalty. 
Here, $x$ denotes the instance being explained, $d$ is the desired class, and $\hat{p}_c(x)$ denotes the predicted probability of class $c$ returned by the black-box model. We minimize $L$, so each component is defined as a non-negative penalty.
We scale $L_{\text{div}},L_{\text{prox}},L_{\text{spars}},L_{\text{plaus}}$ to approximately $[0,1]$ to make the weights $\omega$ easier to interpret, and we use equal weights ($\omega_{\*}=1$) as a neutral default in the benchmark.

The first term ensures the correctness of the counterfactual by enforcing that the predicted probability of the desired class exceeds that of all other classes. 
It is formulated as a hinge loss that becomes zero once this condition is satisfied, guaranteeing the intended classification change. 
In Figure~\ref{fig:case}, this corresponds to moving the query point across the decision boundary. 
However, simply crossing the boundary is not enough to make a counterfactual useful. 
The second term determines where the counterfactual is placed, ensuring it does not end up in an arbitrary region but in a part of the feature space that makes it actionable. 
This is achieved through a weighted combination of property-specific objectives—plausibility, proximity, sparsity, and diversity—that must be carefully balanced. 
For example, a minimal change that violates domain constraints may technically achieve the desired prediction but remains unrealistic and therefore not actionable. 
Figure~\ref{fig:case} illustrates why plausibility is critical: counterfactuals should respect both conditional dependencies (e.g., $P(x_2 \mid x_1, x_3)$ and $P(x_4 \mid x_3)$) and linear relationships among features. 
Ignoring these relationships can lead to invalid suggestions, such as feature combinations that cannot occur in practice. 
By jointly optimizing these properties, the approach ensures that counterfactuals are not only effective in altering predictions but also realistic, interpretable, and suitable for decision-making.
All components are scaled to comparable ranges to allow the use of uniform weights. This provides a neutral default configuration that can be adjusted in application-specific scenarios.

In the following sections, we describe in-depth the design of particular loss components and the process of search for optimal solution.

\subsection{Plausibility}
\begin{figure}[htb]
\centering
\includegraphics[width=\columnwidth]{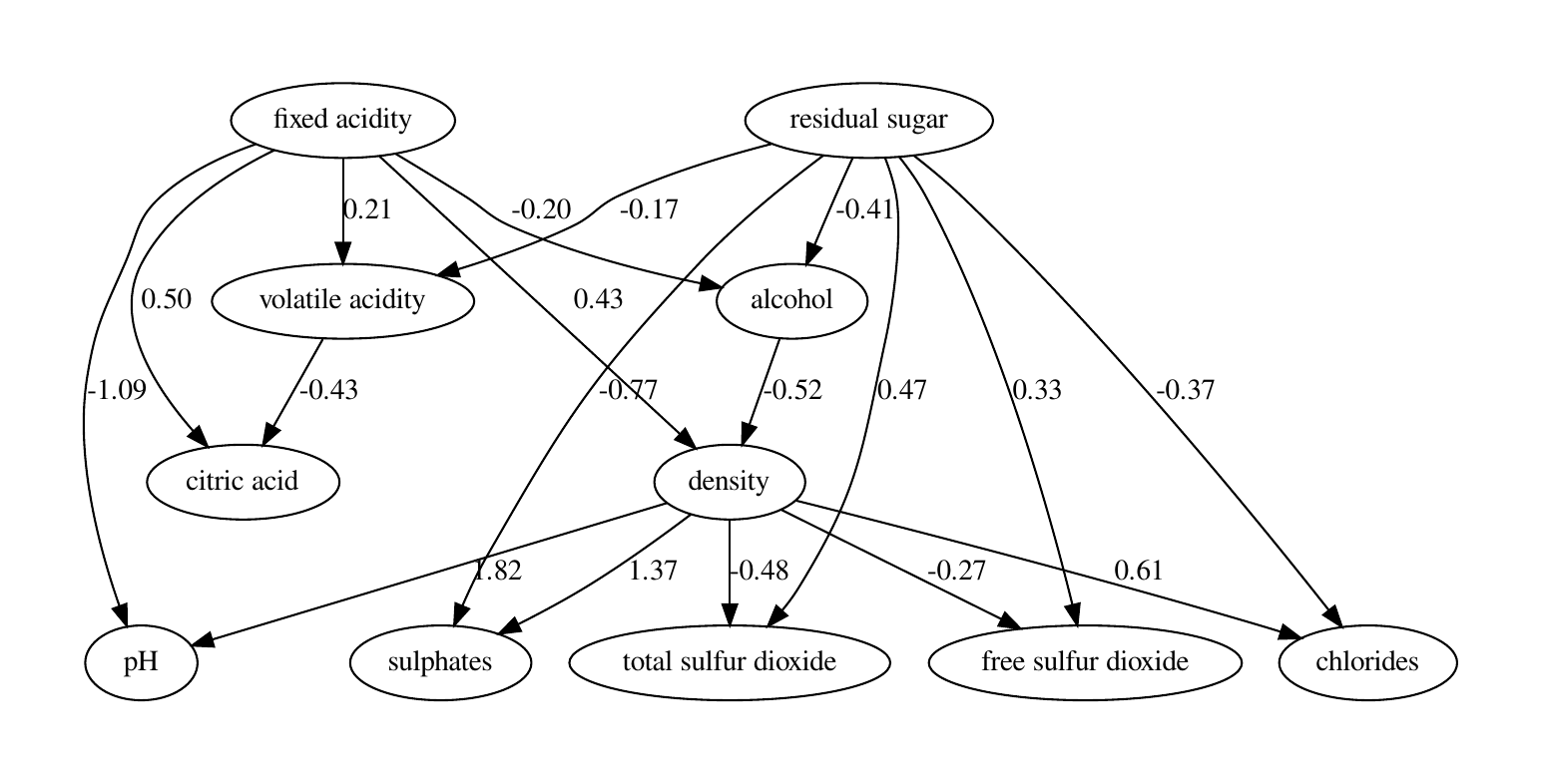}
\caption{Relationship graph for a Wine dataset. The edges represent the magnitude in which change in the parent node values impact the children node values. For instance increase in residual sugar decreases the alcohol in the wine and the counterfactual that does not preserve this constrain is considered implausible and not actionable.}
\label{fig:graph}
\end{figure}

The plausibility in our approach is defined as a consistency of a newly generated sample (counterfactual candidate) with the predefined relationships between feature values.
This relationship can be given a priori by the expert (in which case it may reflect underlying mechanisms) or can be discovered automatically from the data.
In the case of automatic graph discovery, we use the DirectLiNGAM method~\cite{lingam} for linear relationship graph discovery.
For conditional probability graph learning we use again DirectLiNGAM, or alternatively NOTEARS~\cite{notears} for structure learning and estimate probability distribution with Bayesian parameter estimation.
In both cases, the relationships are represented in our approach as a directed acyclic graph (DAG) as presented in Figure~\ref{fig:graph}.
From a practical point of view, in the implementation of our method, the graph is represented as an adjacency matrix.
In the following paragraphs, these two approaches are described in more detail.

\paragraph{Discovery of relationships with graph}
We employ DirectLiNGAM when a linear dependency model is a reasonable approximation and when interpretable edge weights are required, as these can be directly incorporated as soft linear constraints during optimization. As an alternative, NOTEARS provides a continuous optimization framework for learning DAG structures; within our approach, it serves the same purpose of estimating a directed dependency structure and is particularly useful when the assumptions underlying DirectLiNGAM are not satisfied. Importantly, in the large-scale benchmark setting, the learned graph is treated as a structural plausibility model rather than a causal model, and we do not assume a universally valid causal interpretation across heterogeneous datasets.

\noindent\textbf{Expert constraints and conflicts.} When expert-provided constraints are available, we treat them as \emph{hard} structural edges (fixed directions and, if provided, coefficients) and learn only the remaining parts from data. If an expert rule conflicts with data-driven structure, the expert rule takes precedence; contradictory learned edges are removed. Optionally, constraints can be relaxed into soft penalties by increasing $L_{\text{plaus}}$ for violations (future work).

DirectLiNGAM~\cite{lingam} assumes that the data is generated by a linear non-Gaussian acyclic model (LiNGAM), defined as:
\begin{equation}
\mathbf{x} = \mathbf{B} \mathbf{x} + \mathbf{e}
\label{eq:lingam_model}
\end{equation}
where \(\mathbf{x} \in \mathbb{R}^d\) is the vector of observed variables, \(\mathbf{B} \in \mathbb{R}^{d \times d}\) is a strictly lower triangular matrix encoding the directed structure, and \(\mathbf{e} \in \mathbb{R}^d\) is a vector of independent, non-Gaussian noise terms. The method identifies variable ordering by iteratively detecting exogenous variables - those independent of regression residuals. For each variable \(x_j\), residuals \(r_i^{(j)}\) are computed by regressing other variables \(x_i\) on \(x_j\), and independence is tested:
\begin{equation}
x_j \perp r_i^{(j)} \quad \forall i \neq j
\label{eq:lingam_independence}
\end{equation}

The intuition behind this is that if \( x_j \) is truly exogenous, then any dependence of \( x_i \) on \( x_j \) must be due to \( x_j \)'s influence on \( x_i \), not the reverse. 
This is because if we regress \( x_i \) on \( x_j \), the residual \( r_i^{(j)} \) captures the part of \( x_i \) that is not explained by \( x_j \).

If \( x_j \) is exogenous, then it should be statistically independent of the residuals \( r_i^{(j)} \), because \( x_j \) is not influenced by any other variable and
the residuals represent influences from other variables or noise, which are independent of \( x_j \) .

Therefore, if the variable is exogenous, i.e. \(x_j = \mathbf{e_j} \) it becomes the parent node in the graph.
Once the ordering is established, the matrix \(\mathbf{B}\) is estimated via least squares regression.
The \(\mathbf{B}\) represent the linear relationship graph.
For instance, in Figure~\ref{fig:graph} the relationship for well known Wine dataset~\cite{wineds} was presented which was calculated with the described method.
As discussed earlier, the \(\mathbf{B}\)  is strictly lower triangular, therefore no cycles are observed in the graph, and the non-zero elements at the intersection of rows and columns indicate the weight of the edge (the relation between features) depicted in the graph.

\paragraph{Conditional Probability Distribution Estimation}

Once the directed structure is obtained using DirectLiNGAM or NOTEARS, we additionally allow to estimate the conditional probability distributions (CPDs) for each node in the resulting directed acyclic graph (DAG). 
If the observed variables are continuous, we first discretize them into a finite number of bins, transforming the data into categorical form suitable for discrete probabilistic modeling.

We employ the Bayesian parameter estimation, which provides tools for learning CPDs from data given a fixed DAG structure. 
For each variable \( X_i \), the conditional probability distribution \( P(X_i \mid \mathrm{Pa}(X_i)) \) is estimated using maximum likelihood estimation (MLE) or Bayesian estimation, where \( \mathrm{Pa}(X_i) \) denotes the set of parent variables of \( X_i \) in the learned structure.

Each CPD is represented as a conditional probability table that specifies the probability of each discrete state \( X_i = x_i \) given all possible configurations of its parent states \( \mathrm{Pa}(X_i) = \mathbf{x}_{\mathrm{pa}} \):
\[
P(X_i = x_i \mid \mathrm{Pa}(X_i) = \mathbf{x}_{\mathrm{pa}}) = \frac{N(X_i = x_i, \mathrm{Pa}(X_i) = \mathbf{x}_{\mathrm{pa}})}{N(\mathrm{Pa}(X_i) = \mathbf{x}_{\mathrm{pa}})},
\]
where \( N(\cdot) \) denotes the empirical count in the discretized dataset. 

The resulting set of CPDs defines a fully parameterized Bayesian network consistent with the discovered directed structure, enabling probabilistic inference and sampling over the joint distribution of all variables. 
Compared to using the linear coefficients from DirectLiNGAM, this discretized CPD approach has the advantage of capturing potentially nonlinear or non-monotonic relationships between each variable and its parents, since each parent configuration is associated with an empirically estimated probability distribution rather than a fixed linear effect.

%(http://archive.ics.uci.edu/ml/datasets/Auto+MPG) or wine quality

\paragraph{Plausibility loss estimation}
Given a relationship graph $G=(V,E)$, we penalize counterfactual candidates that violate  constraints. We support two realizations of $L_{\text{plaus}}$: Linear dependency mode and probabilistic mode.
%\begin{itemize}
%\item 
In linear dependency mode, each node is reconstructed from its parents using learned edge weights, and we compute a normalized mean squared reconstruction error.
%\item 
In probabilistic dependency mode (used when conditional probability tables are available), we compute a normalized negative log-likelihood under the learned/encoded CPDs, which naturally handles discretized and categorical variables.
%\end{itemize}

In linear mode,  let $\text{cf}_i$ be the $i$-th feature value in the counterfactual candidate and $\mathrm{Pa}(i)$ the parent set of node $i$ in $G$. The reconstructed value is $v_i=\sum_{p\in\mathrm{Pa}(i)} w_{p\to i}\,\text{cf}_p$. The plausibility penalty is defined in Equation~\ref{eq:plausibilityloss} as the average mean squared error between the candidate and its reconstruction, normalized by the number of features $|V|$.
\begin{equation}
 \label{eq:plausibilityloss}
 \begin{aligned}
 L_{\text{plaus}}(\text{cf},G) &= \frac{1}{|V|} \sum_{i\in V} \big(\text{cf}_i - v_i\big)^2,\\
 v_i &= \sum_{p \in \mathrm{Pa}(i)} w_{p\to i}\,\text{cf}_p.
 \end{aligned}
\end{equation}

In probabilistic mode, after discretization the CF instance, for each node $i$ we estimate a CPD $P(X_i\mid\mathrm{Pa}(X_i))$. We then define $L_{\text{plaus}}$ as the average negative log-likelihood of the candidate under these CPDs (optionally normalized), which yields a bounded penalty and supports nonlinear, conditional constraints. The formula is given in Equation~\ref{eq:plausibilityloss_prob}, where $z_i(\text{cf})$ is the discretized value of feature $i$ in the candidate.

\begin{equation}
\label{eq:plausibilityloss_prob}
L_{\text{plaus}}^{\text{prob}}(\text{cf},G)
= \frac{1}{|V|}
\sum_{i\in V}
\left[
-\log \, P\!\left(Z_i=z_i(\text{cf}) \,\middle|\, Z_{\mathrm{Pa}(i)}=z_{\mathrm{Pa}(i)}(\text{cf})\right)
\right].
\end{equation}

\subsection{Proximity}
Proximity encourages counterfactuals to stay close to the explained instance. We use a weighted cosine \emph{distance} on normalized features so that the term can be minimized. Let $\tilde x$ and $\widetilde{\text{cf}}$ denote feature-wise normalized vectors (continuous features scaled to $[0,1]$; categorical one-hot groups treated as blocks). The proximity penalty is defined in Equation~\ref{eq:proximity} as one minus the weighted cosine similarity, where $\omega_i$ are feature-specific weights (e.g., inverse median absolute deviation) that can be tuned to adjust the importance of proximity for different features.
\begin{equation}
 \label{eq:proximity}
 L_{\text{prox}}(\text{cf},x) = 1 - \frac{\sum_{i=1}^{M} \omega_i\,\widetilde{\text{cf}}_i\,\tilde x_i}{\sqrt{\sum_{i=1}^{M} \omega_i\,\widetilde{\text{cf}}_i^2}\;\sqrt{\sum_{i=1}^{M} \omega_i\,\tilde x_i^2}}.
\end{equation}
This quantity is close to $0$ when vectors are aligned.
\subsection{Sparsity}
The sparsity measures how drastic is the modification of an original instance is in order to transform it into counterfactual, which by the definition is the \emph{minimal} change in original feature.
In practice, this is reduced to the number of features that values' were altered in CF compared to original instance, as shown in Equation~\ref{eq:sparsityloss}.

\begin{equation}
    \label{eq:sparsityloss}
    L_{\text{spars}}(\text{cf},x) = \frac{1}{M} \sum_{i=1}^{M} \mathbb{I} (\text{cf}_i \neq x_{i})
\end{equation}

Note, that the function is non-convex, hence it cannot be included in the loss function if the optimization is gradient-based.
However, but we use optimization method that works with non-convex functions and allows us to include it in the total loss.

\subsection{Diversity}
To encourage a set of \emph{distinct} counterfactuals, we follow the determinantal point process (DPP) style objective used in~\cite{dice}. We define a similarity kernel matrix $K$ over the already selected counterfactuals $CF=\{\text{cf}^{(1)},\ldots,\text{cf}^{(N)}\}$ and minimize a bounded penalty as shown in Equation~\ref{eq:diversityloss}, where $\det(K)$ is the determinant of $K$ and serves as a measure of diversity (higher determinant indicates more diverse set).
\begin{equation}
 \label{eq:diversityloss}
 L_{\text{div}}(CF) = 1 - \det(K).
\end{equation}
We use a kernel based on Euclidean distance in the normalized space:
$$
K_{ij} = \frac{1}{1 + \lVert\widetilde{\text{cf}}^{(i)}-\widetilde{\text{cf}}^{(j)}\rVert_2} + \epsilon\,\delta_{ij},
$$
where $\delta_{ij}$ is the Kronecker delta and $\epsilon$ is a small constant (we use $10^{-6}$) that stabilizes the determinant by ensuring numerical well-conditioning. With this construction $\det(K)\in(0,1]$, so $L_{\text{div}}\in[0,1)$.

All of the components $L_{\text{div}},L_{\text{prox}},L_{\text{spars}},L_{\text{plaus}}$ described before are finally added to the global loss function presented in Equation~\ref{eq:ccfloss} and optimized with an approach presented in the following section.
%The $L_d,L_n,L_s,L_p$ described before are finally added to the global loss function presented in Equation~\ref{eq:ccfloss} and optimized with an approach presented in following section.

\subsection{Counterfactual search}
%Introduce parzen search, search space initialization (linear models, train set, and plausibility model) sampling.
The process of search for counterfactuals can be divided into three stages that are executed in the following order. 
The compact representation of this process is also presented in Algorithm~\ref{alg:ccf}.
\begin{enumerate}

\item In the first stage, the relationship graph is created. It can be either generated automatically in a data-driven manner or provided explicitly by the expert. If the full graph is not known, only partial relationships can be defined. 

\item In the second stage, data generation is performed to create a pool of counterfactual candidates. The initial set of candidates is generated by sampling from the provided dataset, or, if unavailable, by generative sampling from the relationship graph. In the case of additive and linear models, data can also be generated using information about feature contributions similar to~\cite{dce} and~\cite{dace}.

\item Finally, the loss function is optimized using a Tree-Structured Parzen Estimator~\cite{parzen} which is a kind of Bayesian optimization approach that allows for tree structure search spaces (i.e. spaces with possible conditional parameters).
\end{enumerate}
\noindent\textbf{Search space and initialization.} The search space is defined per feature by (i) empirical min--max ranges from the training data, (ii) optional expert-provided bounds and immutability masks (features that must not change), and (iii) relationship-graph-based feasibility filtering (linear or probabilistic mode). Initial candidates $X'$ are drawn from (a) observed samples (including nearest opposite-class samples when available) and (b) ancestral sampling from the relationship graph when enabled. The Tree-structured Parzen Estimator then proposes candidate edits within these bounds, and candidates that violate hard constraints are rejected while soft violations are penalized by $L_{\text{plaus}}$.

\begin{algorithm}[tb]
   \caption{DANCE Counterfactual search algorithm}
   \label{alg:ccf}
\begin{algorithmic}[1]
   \State {\bfseries Input:} Blackbox Model $M$; Training dataset $X$; Relationship graph $G$; Desired number of counterfactuals $N$; Weights $\omega$ for the loss function from Equation~(\ref{eq:ccfloss}); Desired class $d$
   \State {\bfseries Output:} List $CF$  of $N$ counterfactual examples
   \State
      \State Initialize result set $CF$ \assign $\emptyset$;
   \If{$G = \emptyset$}
        \State Train $G$ with $X$ and LiNGAM, or Bayesian parameter estimation
   \EndIf
   \For{$i \in  1,2,\ldots,N$ }
        \State Create initial counterfactual candidates $X'$
        \State $X'$ \assign samples from $X$ and generated with $G$
        \If{$M$ is additive or linear}
            \State $X'$ \assign samples generated from $M$
        \EndIf
        \State Find counterfactual $cf_i \assign \arg\min_{cf_i} L(M, cf_i, CF, G, X',d, \omega)$
        \State Add $cf_i$ to the result set $CF$ \assign $\{cf_i\}  \cup CF$
   \EndFor
   \State {\bfseries Return:} List of plausible, diverse, and sparse counterfactuals $CF$
\end{algorithmic}
\end{algorithm}

%FIXME ja mam ogolnie duzy problem z tym jak i gdzie jest mowa o tym grafie i jego budowie. Najpierw bylo we wstepie ale bez nadmiaru komentarza (ktory przenioslem do metody), w metodzie odnosimy sie wczesnie do tych pojec, po czym na KONCU opisu metody piszemy, ze to jest core part. Ja bym dobrze przemyslal te narracje. i pytanie czy to co jest nizej nie powinno byc omowione wczesniej

One of the most important parts of the method is the relationship graph that captures directed dependencies between variables and represents linear or probabilistic relationships between features' values.
Figure~\ref{fig:graph} presents a sample graph trained on the well-known Wine dataset~\cite{wineds} which was selected solely for illustrative purposes due to its simplicity and widespread familiarity, which help to clearly demonstrate the underlying concepts.
The structure and coefficients on the edges were learned using DirectLiNGAM approach. 
As mentioned previously, the graph paragraph can be obtained from experts (fully or only partially) or can be learned automatically from the data.
If the graph is provided by the expert, it does not have to be complete, as the method automatically finds missing relationships from the data.
The graph is used in the plausibility loss component to  ensure that the counterfactual is aligned with it and does not contain feature values that are not consistent with the background knowledge.

Additionally to relationship graph, we can narrow the search space with range-based constraints, such that prevent certain variables from taking certain values (i.e. values from outside of the domain, or forbidden values (e.g. age of a person lower than actual). 
Such constraints also allow us to mask features that should not be modified at all by the algorithm.

Our method ensures diversity among counterfactuals, allowing the desired number of counterfactuals to be specified as output. 
Counterfactuals are generated sequentially: in the first iteration, the best counterfactual is identified, and in subsequent iterations, additional diverse counterfactuals are sought.

\section{Quantitative evaluation}
\label{sec:eval}

%TODO: more details on the setup, how graph is derived, technical stuff related to hyperparameters choice, etc.

We compared our method with selected state-of-the-art approaches, focusing exclusively on model-agnostic methods for which publicly available implementations exist. 
In addition to the full version of our approach, we conducted an ablation study in which the plausibility component was removed in order to assess its influence on the remaining criteria, such as diversity and sparsity. 
All experiments were conducted on 140 datasets from the OpenML repository\footnote{See: \url{https://www.openml.org/}.}
, using Explainable Boosting Machines (EBM)~\cite{caruana2013ebm} as the underlying classification model. We chose EBM for its stable probabilistic outputs and availability across all datasets; however, the DANCE optimizer only queries $\hat p_c(\cdot)$ and does not rely on gradients or internal model structure, so it is model-agnostic by design. 
Due to the large number and heterogeneity of datasets considered, expert-defined relationship graphs were not feasible; therefore, relationship graphs were automatically discovered from data using DirectLiNGAM. 
Expert-validated graphs were used only in the dedicated use-case scenario described later in the paper (Section~\ref{sec:userstudy}). 
Unless stated otherwise, all weights $\omega_{\*}$ in Eq.~\ref{eq:ccfloss} were set to 1.0. Because each loss component is scaled to a comparable range, equal weights form a neutral default that avoids ad-hoc tuning across 140 heterogeneous datasets. We discuss this choice and its implications in Section~\ref{sec:limitations}. 
To guarantee a fair comparison, each evaluated method was provided with exactly the same instances for counterfactual explanation and was instructed to generate three diverse counterfactuals per instance. 
For methods that do not natively support diversity, the explanation process was repeated three times consecutively. 
All evaluation metrics were computed strictly according to the definitions and equations presented in Section~\ref{sec:bkg}. 
Detailed quantitative results for individual properties are reported in the appendices. The source code required to reproduce all experiments is publicly available in our GitHub repository\footnote{See: \url{https://anonymous.4open.science/r/xaibench-ACD8/}.}, and precomputed metrics for all methods and datasets are provided as supplementary material.

%TODO: more one the evaluation setup: where the coasal graph were taken from, how the hyperparameters were chosen--i.e. the weights for the equaiton, etc.

\begin{table*}
\fontsize{9}{10.8}\selectfont
\caption{Summary of results from quantitative evaluation. Values represent average rank of the method over 140 datasets. The lower the better. The $\uparrow$ denotes winners that were statistically significantly better the the rest.}
\label{tab:summaryrank}
\begin{tabularx}{\textwidth}{p{2.5cm}XXXXXXXX}
\hline
 & DANCE&Dice&LUX&LORE&CEM&CEGP&Wachter&cfnow
\\ \hline \hline
Fidelity & \textbf{2.86} & 3.54 & 4.16 & 5.74 & 5.89 & 4.21 & 5.98 & 3.61\\ \hline
Probability & \textbf{1.75} $\uparrow$ & 3.87 & 4.31 & 4.83 & 5.81 & 5.47 & 5.62 & 4.34\\ \hline
Proximity Loss & \textbf{2.55} & 4.01 & 4.36 & 5.88 & 5.81 & 3.91 & 6.17 & 3.29\\ \hline
Sparsity Loss & \textbf{2.03} $\uparrow$ & 3.21 & 5.03 & 6.19 & 5.12 & 4.54 & 6.48 & 3.40\\ \hline
Plausibility Loss & \textbf{2.54} $\uparrow$& 4.33 & 3.55 & 5.21 & 6.14 & 4.24 & 5.92 & 4.06\\ \hline
Diversity & 4.34 & 3.44 & 4.92 & 3.92 & 4.88 & 4.57 & \textbf{2.96} & 6.97\\ \hline
Execution Time & \textbf{2.37} & 3.67 & 3.14 & 5.04 & 6.21 & 5.11 & 6.34 & 4.12\\ \hline
Total Wins & \textbf{6} & 0 & 0 & 0 & 0 & 0 & 1 & 0\\ \hline
\end{tabularx}
\end{table*}

\subsection{Analysis of the results}

\begin{figure}[htb]
\centering
\includegraphics[width=0.9\columnwidth]{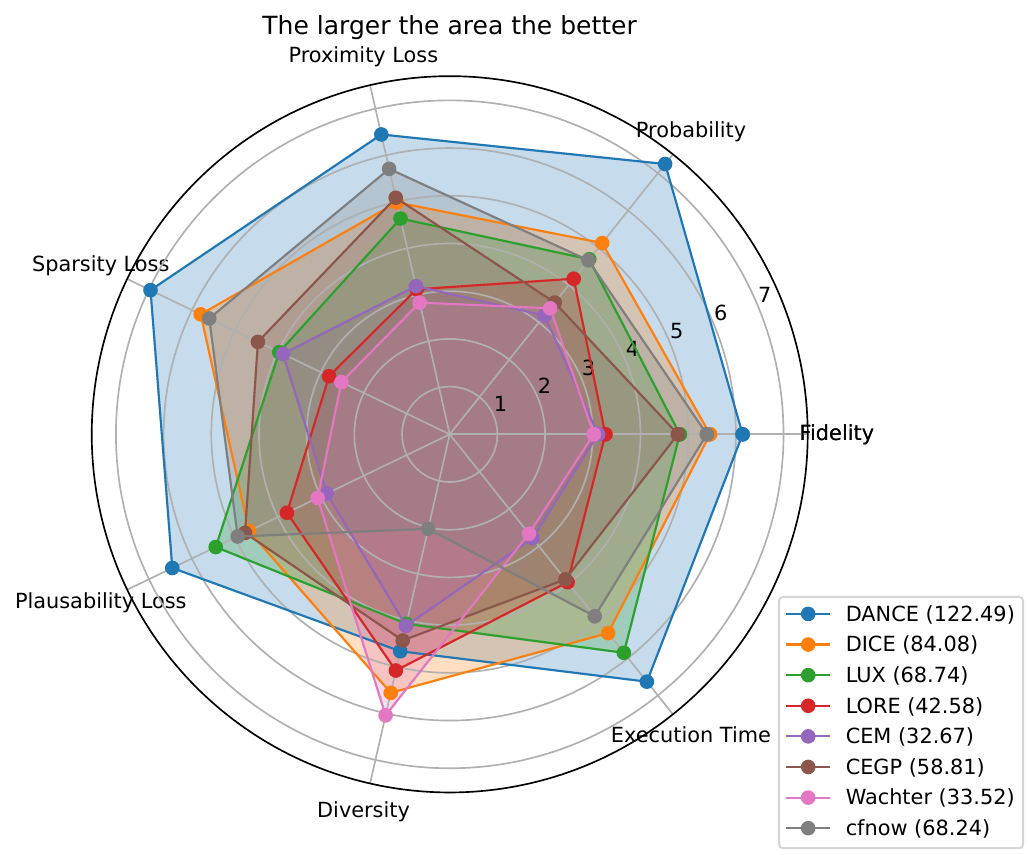}
\caption{Combined evaluation metric visualized in a form of spiderplot. The larger the area, the better the method.}
\label{fig:spider}
\end{figure}

%TODO provide detailed analysis in appendices
A summary of these results is shown in Figure~\ref{fig:spider}. 
The plot presents the scores obtained by each method across 140 datasets. 
The score is calculated as the area under a spider plot (AUS) in a form of sum areas of triangles defined by the center of the plot and neighborhood vertices. 
Each vertex representing the $\frac{1}{R_i}$ of a method, where $R_i$ is the average ranking obtained from Friedman post-hoc statistical tests for the $i-th$ property of a given method. 
The formulas for this combined metric is given in Equation~(\ref{eq:spider}).
The closer a method's score is to the maximum possible area defined as $\frac{NM^2}{2}  \sin\left(\frac{2\pi}{N}\right)$, the better its overall performance. The $M$ represents the total number of methods tested.

\begin{equation}
\label{eq:spider}
AUS = \frac{1}{2} \sin\left(\frac{2\pi}{N}\right) \sum_{i=1}^{N} \frac{1}{R_i R_{i+1}}
\end{equation}

This evaluation method is adopted from~\cite{lux}, where the authors use it to illustrate multi-criteria evaluation results in scenarios where there is a known trade-off between the metrics.
The area can be affected by the order of vertices, hence it is recommended to put metrics that there is known trade-off between (e.g. precision and recall) next to each other. 
In the plot, we included \textit{probability} metric showing how confident the model predictions were for a given counterfactual.
In our case, all metrics are treated equally as the trade-off exists between all of them.

Additionally, we tested the statistical significance of the differences of our method compared to the rest.
We have conducted Friedman post-hoc test followed by Nemenyi test and have proven that our method is significantly better than others in plausibility and sparsity and probability metrics.
In the remaining metrics, it also achieved top scores, but the differences were not statistically significant.
The only metric that our method was not the best was diversity, where it lost to Wachter, DICE and LORE. 
 This is particularly remarkable because Wachter and LORE does not inherently optimize counterfactuals for diversity. 
 Instead, we simulated this by running Wachter and LORE multiple times and collecting results from each run. 
 As a result, the observed diversity was primarily due to the stochastic nature of counterfactual generation implemented within these methods.
The critical distance plots and statistical analysis of the tests are provided in~\ref{ap:test}.

\section{Statistical tests for quantitative evaluation}
\label{ap:test}
With 8 algorithms and 140 datasets, we had 7 x 973 degrees of freedom, respectively.
This enabled us to establish that the critical distance for the post-hoc Nemenyi test, with  $F( 7 , 973 )$ and $\alpha=0.05$ is 0.75 for all the subsequent cases.

\begin{figure}[htb]
\centering
\includegraphics[width=\columnwidth]{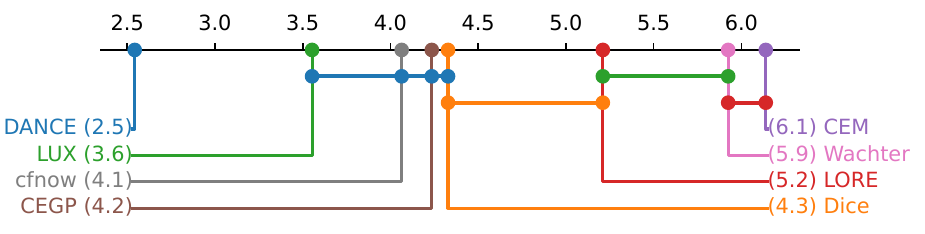}
\caption{Critical distance plot for plausibility loss.}
\label{fig:plausability}
\end{figure}

Figure~\ref{fig:plausability} shows that DANCE achieves the highest plausibility (Eq.~\ref{eq:plausibilityloss}) with statistically significant differences. Occasional inferior results to LUX stem from its endogenous nature, selecting only existing (thus always plausible) instances.

\begin{figure}[htb]
\centering
\includegraphics[width=\columnwidth]{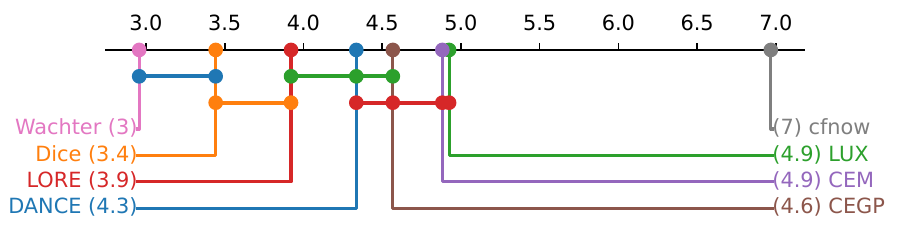}
\caption{Critical distance plot for diversity.}
\label{fig:diversity}
\end{figure}

Figure~\ref{fig:diversity} presents the results DANCE achieved in terms of diversity metric defined in Equation~(\ref{eq:diversityloss}).
It is notable that the results were comparable to LORE and CEGP, but is statistically worse than Dice and Wachter.
It can be observed that the diversity comes with a trade off to plausibility, by comparing Figure~\ref{fig:diversity}  and Figure~\ref{fig:plausability}. This is especially visible in case of Wachter, which took opposite places for both of these metrics. 
In many cases assuring plausibility comes at a cost of diversity, especially where there are strong relationships between variables, that prevent diverse changes to their values.

\begin{figure}[htb]
\centering
\includegraphics[width=\columnwidth]{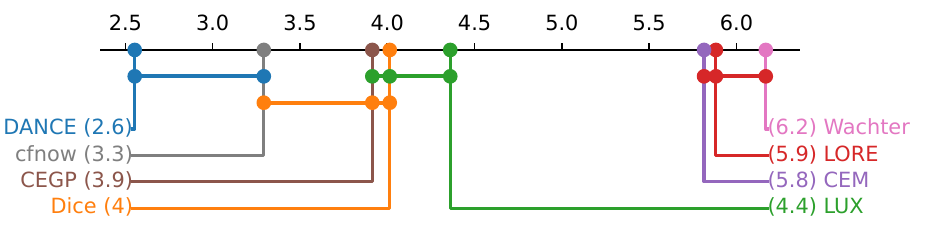}
\caption{Critical distance plot for proximity loss.}
\label{fig:proximity}
\end{figure}

\begin{figure}[htb]
\centering
\includegraphics[width=\columnwidth]{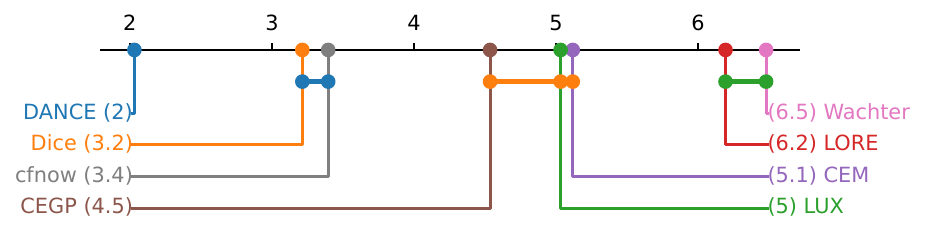}
\caption{Critical distance plot for sparsity loss.}
\label{fig:sparsity}
\end{figure}

Figure~\ref{fig:proximity} shows the results of the proximity evaluation defined in Equation~(\ref{eq:proximity}).
The DANCE obtained the best results along with CFNOW. Proximity correlates with sparsity (Fig.~\ref{fig:sparsity}, Eq.~\ref{eq:sparsityloss}), with sparse methods achieving better proximity.

\begin{figure}[htb]
\centering
\includegraphics[width=\columnwidth]{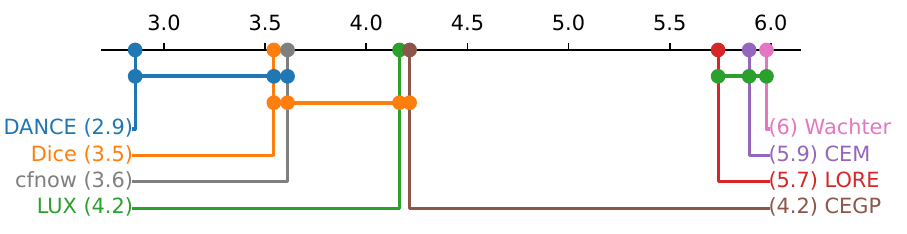}
\caption{Critical distance plot for fidelity.}
\label{fig:fidelity}
\end{figure}

Figure~\ref{fig:fidelity} presents the results for the fidelity metric that measures ratio of correctly discovered counterfactuals (counterfactulas that indeed changed the prediction of the model).
A clearly visible two clusters of methods were formed here with a significant gap between them.

\begin{figure}[htb]
\centering
\includegraphics[width=\columnwidth]{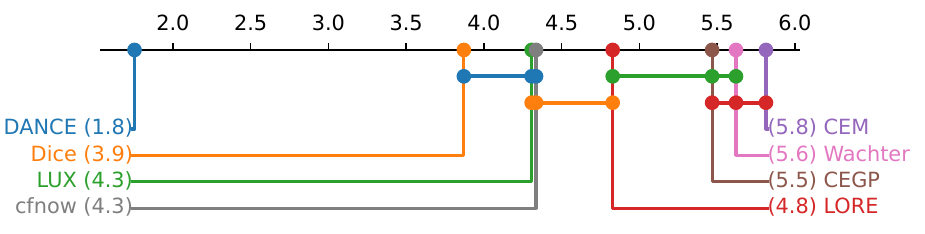}
\caption{Critical distance plot for probability.}
\label{fig:probability}
\end{figure}

Figure~\ref{fig:probability} shows the results obtained for the probability metric.
It is highly correlated with fidelity, as it shows the probability of the counterfactual obtained from the black-box model for a desired class.
However, in this plot it can be seen that although the fidelity metric was similar to several algorithms, in case of probability metric, the DANCE stands out as the best one.

\begin{figure}[htb]
\centering
\includegraphics[width=\columnwidth]{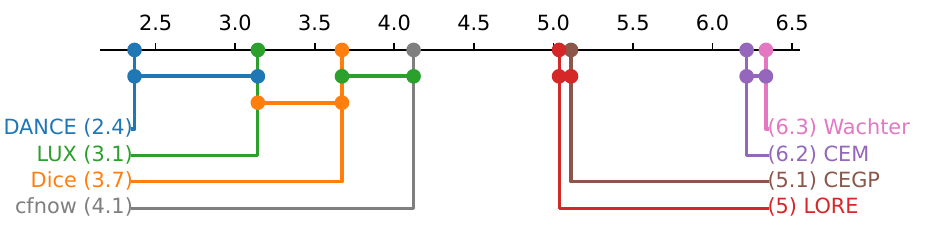}
\caption{Critical distance plot for execution time.}
\label{fig:execution}
\end{figure}

The evaluation of execution time was presented in Figure~\ref{fig:execution}.
%It is worth noting that only the execution time of the search for counterfactuals were measures in the case of DANCE. 
%The creation of a relationship graph was not included in the measurements.
The reported execution time includes only the counterfactual search phase. The construction of the relationship graph is treated as an offline preprocessing step and is therefore excluded from runtime comparisons, as it can be reused across multiple explanation queries.
%The creation of a relationship graph is very time consuming, but it is a one-time procedure that can be combined with balck-box model training, hence we excluded it from the evaluation.

\subsection{Ablation study}
We restrict the ablation analysis to removing the plausibility component because enforcing structured dependencies is the primary additional mechanism introduced by DANCE; the remaining terms follow standard formulations used in prior work and are included to provide a balanced objective.

In this study, we removed the plausibility objective from the total loss function to observe how it affects the overall performance of other properties.
In particular, we aimed at finding if plausibility negatively affects other metrics, especially such as diversity or sparsity.
We tested these two models on the same set of datasets as discussed in previous section.

\begin{figure}[htb]
\centering
\includegraphics[width=0.9\columnwidth]{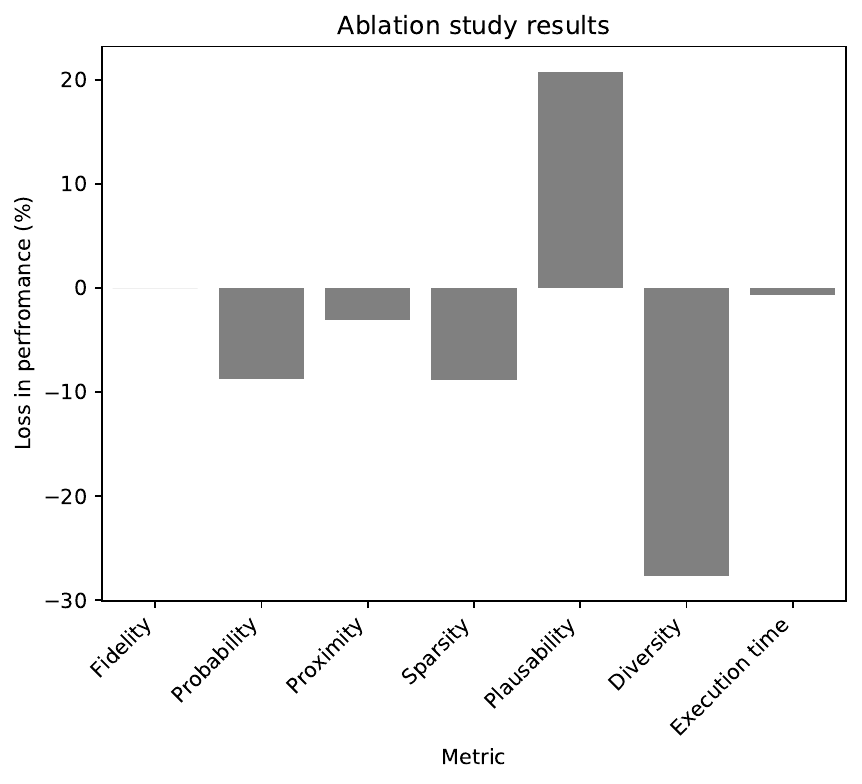}
\caption{Decrease in performance with respect to different metrics of a model with plausibility optimization turned on.}
\label{fig:ablation}
\end{figure}

The difference in performance of the model with plausibility turned on relative to the model, where we turned off the plausibility component, is presented in Figure~\ref{fig:ablation}.
The most noticeable difference is observed in plausibility, which is approximately 20\% higher for the model incorporating the plausibility component.
Conversely, the metrics visibly negatively impacted are diversity and  sparsity.
Also, the probability of the predictions of the black-box model for generated counterfactuals dropped by 10\%  in plausible model.
The fidelity remains the same for both solutions with small changes to proximity and execution time.

The reduction in sparsity and diversity arises from the method's inherently constrained search space for counterfactuals, guided by the relationship graph.
This constraint leads to cascading modifications of dependent features within the graph, thereby reducing sparsity.
Similarly, the diversity is constrained by the relationship graph, as there are areas of the feature space that cannot be explored without a negative impact on plausibility.

\section{Evaluation in the business case study}
\label{sec:userstudy}
In this section we present a case study of the practical usage of DANCE method in real business case.
The case study is a part of a Sendguard project carried out in collaboration with Freshmail company that one of the main operating goals is delivering tools that allow clients to design, manage, and send their own marketing campaigns through CPaaS approach to guard their quality.
Despite adhering to best practices and maintaining a strong emphasis on quality and consent, or due to the security challenges, the company mass distribution of emails frequently faces the risk of being misclassified as spam. 

% However, the ultimate goal of the project  goes beyond simply labeling campaigns as \emph{spam} or \emph{ham} messages. 
% It incorporates an actionability requirement, enabling experts to modify campaigns in ways that enhance their quality and boost metrics such as the open rate, a key indicator of user engagement. 
% Non-actionable recommendations would hold little value in a business scenario, as they fail to provide practical guidance for improvement.

The core challenge driving this study stems from the observation that some clients of company's CPaaS platform, despite developing ostensibly high-quality email campaigns (e.g., regularly updated subscriber lists, thematically relevant and appealing content), often achieve suboptimal user engagement rates. 
Analyzes revealed that issues such as poorly constructed subject lines or suboptimal sending times can overshadow otherwise robust campaign content. 
Consequently, these campaigns frequently yield results similar to, or even worse than, campaigns of objectively lower content quality (e.g., stale subscriber lists, purely promotional or repetitive messages).

\begin{figure*}[htb]
\centering
\includegraphics[width=0.9\textwidth]{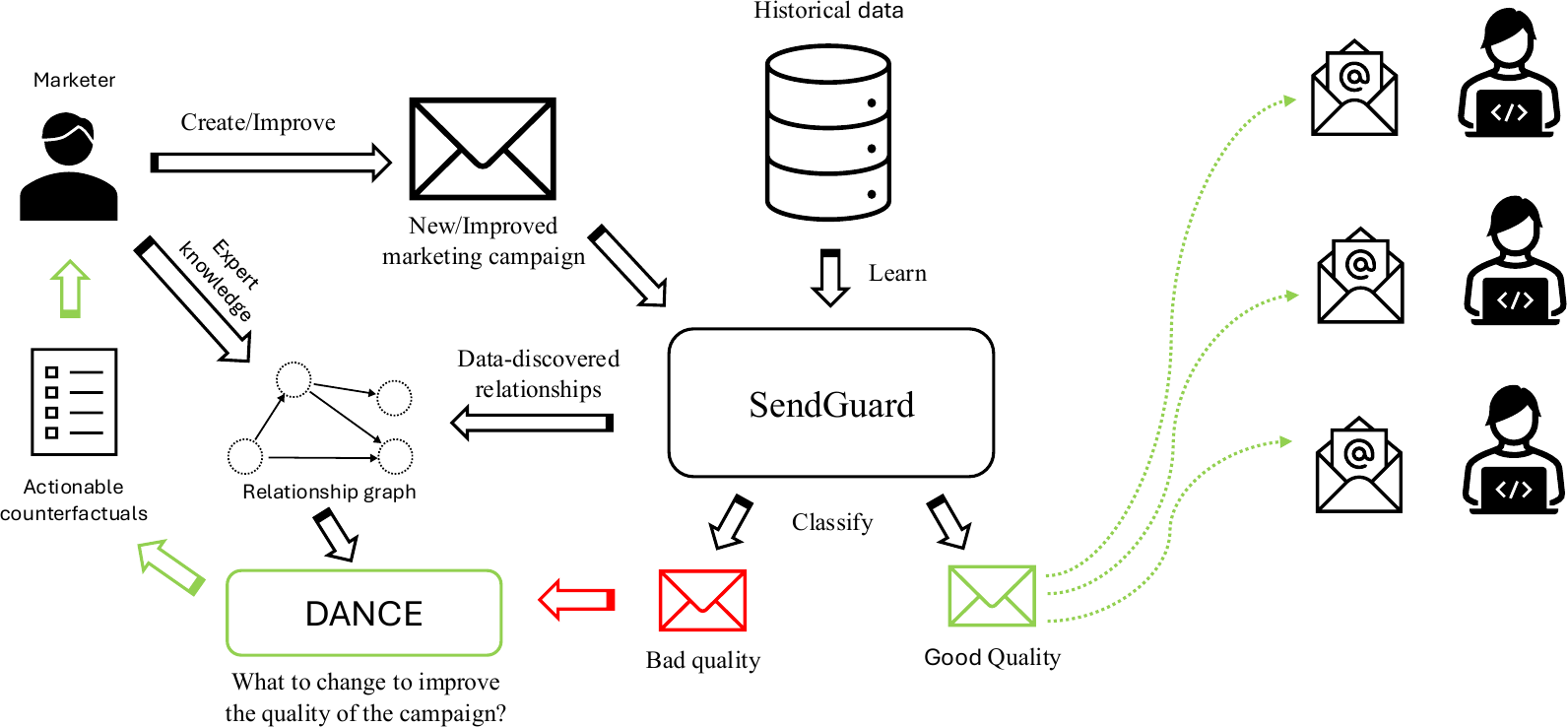}
\caption{Workflow of the case study. The marketing campaign is processed through the Sendguard system to assess its quality. If the campaign is deemed of poor quality, the DANCE method generates counterfactual suggestions for potential modifications to enhance its possible open rate among addressees.}
\label{fig:workflow}
\end{figure*}

In light of this, our primary objective was to devise a mechanism that suggests optimizations of selected campaign parameters, particularly those most visible to recipients (e.g., subject line elements, sending hour) and most likely to affect open and click rates. 
Rather than relying solely on heuristic guidelines, we sought a data-driven approach that would integrate expert knowledge about email marketing practices while respecting the intricate dependencies among features.
In this case, the counterfactuals delivered by DANCE are a solution to our case.
The overall idea of the system incorporating DANCE method is presented in Figure~\ref{fig:workflow}.
In the following sections we elaborate on the dataset we used and results obtained with our approach.

% Moreover, while generating counterfactual explanations, it is crucial to preserve well-established good practices and known relationships between campaign properties. 
% This ensures that any recommendations align with industry standards and do not inadvertently degrade the campaign's effectiveness or credibility.

\subsection{Dataset description}

In our study we used company's own dataset spanning the period of 
2019 to 2021. 
Part of this dataset was made publicly available on the Zenodo platform~\cite{sbk2025sg900k}].
The total number of analyzed campaigns exceeded one million, though for model building we utilized a stratified sample of 100,000 campaigns (66\% for training, 33\% for testing). 
This sample captured a broad cross-section of sending behaviors, content styles, and thematic tags.
Importantly, the dataset does not include any personal or sensitive information. 
By design, only the most influential high-level parameters were retained, thereby ensuring broad applicability across different types of email campaigns. 
This not only avoids potential biases or discrimination in feature selection but also makes the modeling approach scalable regardless of individual subscriber histories or sensitive demographic data.
Each campaign record was described by the following primary features:
\begin{itemize}
\item Tag (e.g., Sport, Heavy-Industry, Events, Toys, etc.): Captures the general content category.
\item Sending Hour and Day of the Week: Indicate the temporal configuration of the campaign.
\item Content Composition: Includes the number of links, the number of images, and a text-to-HTML ratio measure.
\item Subject Line Properties: Subject length, presence of personalization tokens, emotive language, or question marks.
\end{itemize}

The target variable identified whether a campaign was considered \emph{bad quality} (0) or \emph{neutral/good quality} (1). 
These classes were derived from aggregate open rates, click rates, spam complaint rates, and unsubscribe rates, combined into a weighted score. 
This approach allowed us to separate lower-performing campaigns from those reaching acceptable or positive engagement benchmarks.

For the core predictive model, we employed the Explainable Boosting Machine (EBM) from the InterpretML toolkit achieving the following evaluation metrics:
Accuracy: 0.6707, Recall: 0.6707, Precision: 0.6724, F1 Score: 0.6427, ROC AUC: 0.7471.
Our primary aim was not to maximize predictive accuracy at all costs. 
Instead, we required a sufficiently robust classifier that could differentiate between suboptimal and acceptable campaigns while remaining interpretable. 
The ROC AUC of approximately 0.75 satisfied our baseline requirement for deployment and subsequent counterfactual analysis.

\subsection{Relationship graph}

\begin{figure}[htb]
\centering
\includegraphics[width=1\columnwidth]{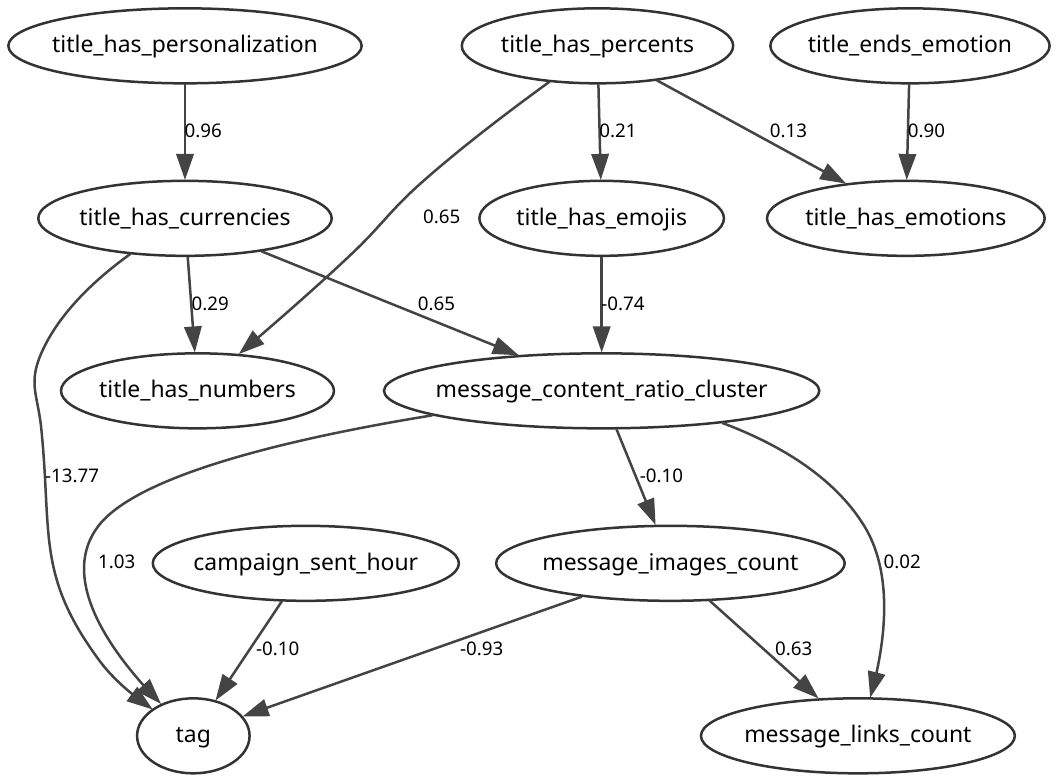}
\caption{Relationship graph for the business case, created in collaboration in domain experts from the marketing field.}
\label{fig:frm-rg}
\end{figure}

A key premise of our method is that the optimal campaign parameters must reflect established best practices and domain-specific insights. 
For example, subject lines containing numbers (e.g., "5 Ways to...") or percentages (e.g., "Save 20\%") often draw attention and bolster open rates. 
Conversely, overuse of emotional triggers or special symbols can trigger spam filters or frustrate recipients. 
Similarly, it is widely acknowledged that sending hours can significantly affect campaign performance, as recipients in certain industries are more likely to engage with emails during specific windows (e.g. early morning for B2B segments).
Through close collaboration with marketing experts, we formalized these observations as a set of business rules governing how features such as \verb$title_has_numbers$, \verb$title_has_emotions$, \verb$title_has_emojis$, and \verb$campaign_sent_hour$ should interplay. 
The business rules were then encoded in the relationship graph used by the DANCE method and are depicted in Figure~\ref{fig:frm-rg}.
Notably, the brand context (B2B vs. B2C) and content category (tag) further modulate which strategies are likely to resonate with different audiences. 
This expert knowledge prevents the optimization process from suggesting unrealistic or domain-inconsistent changes—thereby ensuring the feasibility and credibility of any recommended modifications.

\subsection{Counterfactual generation}

To translate the above insights into actionable changes, we used DANCE %developed a Counterfactual Generation mechanism (referred to here as DANCE)
that integrates both the learned model (EBM) and expert-driven constraints. 
%While detailed explanations of the structural assumptions (e.g., via DirectLiNGAM) are presented in a separate section, the essence lies in respecting known relationships between features during counterfactual construction. 
This prevents suggestions that violate fundamental marketing principles or appear unrealistic in real-world deployments.
In keeping with our initial design premise - to optimize only those parameters that recipients directly perceive before deciding whether to open a message or mark it as spam -- we deliberately excluded from the set of optimizable features any that pertain explicitly to the content domain (tag) or to the overall message composition features.
The rationale behind this constraint is twofold: first, thematic or domain-specific elements often require more substantial, brand-aligned revisions than simple parameter tweaks; second, altering deeper structural aspects of a campaign's body (e.g., message layout, multimedia elements) goes beyond the minimal, high-impact changes that our mechanism aims to propose.
In practice, the counterfactual generator identifies campaigns predicted to be \emph{bad quality} (0) and proposes minimal yet meaningful feature shifts-such as modifying sending hour or introducing certain elements in the subject line-that could alter the model's prediction to \emph{neutral/good quality} (1). It does so by:

\begin{enumerate}
\item Checking feasible ranges and valid combinations for each allowable feature (e.g., limiting emotive punctuation in B2B contexts).
\item Prioritizing changes that are easiest to implement and most likely to improve engagement (e.g., adjusting \verb|campaign_sent_hour|  before attempting more drastic subject line modifications).
\item Maintaining a record of which features were changed and by how much, allowing marketers to evaluate both the acceptability and practical effort of each proposed adjustment.
\end{enumerate}

\subsection{Results and discussion}

We tested our approach on a holdout set of 5,000 campaigns initially classified as \emph{bad quality}. The counterfactual generator was able to propose at least one successful modification (i.e., one that changed the predicted class to \emph{neutral/good quality}) for 3,248 samples, corresponding to 64.96\% of this subset.

The sending hour (\verb|campaign_sent_hour|) was by far the most common parameter to change (4,455 times), suggesting that a poorly timed send can markedly reduce engagement. 
The next most frequent changes involved the presence of emojis, numbers, and percentages in the subject line. 

Additionally, 4,351 of the generated counterfactuals required exactly one feature change, while 732 involved two features. 
Only 107 cases required three or more modifications. 
These findings confirm that even relatively small tweaks -- especially in timing -- can often suffice to transition a campaign from low-performing to adequate or better.
The exemplary explanation of a sample campaign with domain expert comment regarding it was presented in Table~\ref{tab:frm-example}.

\begin{table}[h!]
\centering
\caption{Comparison of feature values between sample campaign marked as \emph{bad quality} and counterfactual generated with DANCE. Bold rows indicate values that were changed by the mechanism.}
\label{tab:frm-example}
\begin{tabular}{lll}%{\textwidth}{XXX}
\hline
\textbf{Campaign feature} & \textbf{Origin} & \textbf{DANCE} \\ \hline\hline
title\_has\_numbers & 1 & 1 \\ \hline
title\_has\_personalization & 0 & 0 \\ \hline
\textbf{title\_has\_percents} & \textbf{1} & \textbf{0} \\ \hline
title\_has\_emotions & 1 & 1 \\ \hline
title\_has\_questions & 0 & 0 \\ \hline
title\_has\_quotes & 0 & 0 \\ \hline
\textbf{title\_has\_emojis} & \textbf{1} & \textbf{0} \\ \hline
title\_has\_currencies & 0 & 0 \\ \hline
title\_has\_brackets & 0 & 0 \\ \hline
title\_ends\_question & 0 & 0 \\ \hline
title\_ends\_emotion & 1 & 1 \\ \hline
\textbf{campaign\_sent\_hour} & \textbf{11} & \textbf{8} \\ \hline
campaign\_sent\_dayofweek & 5 & 5 \\ \hline
campaign\_sent\_day & 28 & 28 \\ \hline
tag & Cosmetic, Beauty & Cosmetic, Beauty \\ \hline
message\_content\_ratio\_cluster & 1 & 1 \\ \hline
message\_links\_count & 24.0 & 24.0 \\ \hline
message\_images\_count & 14.0 & 14.0 \\ \hline
\end{tabular}
\end{table}

In the illustrated example in Table~\ref{tab:frm-example}, the DANCE mechanism proposed modifications to the subject line by removing percentages and emojis. 
This combination is commonly encountered in promotional campaigns within the "Cosmetics and Beauty" domain, where messages often highlight discounts or special offers. 
However, such content frequently receives a lukewarm reception and can even be perceived negatively. 
Consequently, the recommended adjustment encourages employing alternative content types rather than relying primarily on explicit promotional cues.
Moreover, the mechanism advised changing the dispatch time from 11:00 AM to 8:00 AM. 
Empirical observations show that this audience segment is more inclined to engage with such content during morning inbox checks; messages sent later in the workday are frequently deprioritized and overlooked as lower-importance communications.

To further verify that domain constraints were respected, we grouped campaigns by tag (B2B, B2C, Mixed) and examined the statistical distribution of proposed feature values. Notably, the distribution of \verb|campaign_sent_hour| in post-optimization campaigns varied significantly across these segments, as indicated by a chi-square test ( $\chi^2 = 104.45$, and $p < 0.000001$), suggesting the mechanism indeed adjusted sending times in ways consistent with the distinct preferences of different audience groups. 
This alignment with established industry practices underscores the value of incorporating domain expertise into the optimization logic, rather than relying on purely data-driven heuristics.

\section{Limitations}
\label{sec:limitations}
Several limitations affect interpretation. In the OpenML benchmark we learn a directed relationship graph from observational data and use it as a structural constraint in counterfactual search. Because such graphs rely on strong assumptions (e.g., linearity, non-Gaussian noise, and acyclicity for DirectLiNGAM) that may not hold across heterogeneous datasets, results should be read as plausibility under a \emph{learned plausibility model}, not as universally valid causal directions. As the learned graph is approximate, missing/spurious edges or wrong directions can make constraints too permissive (admitting implausible counterfactuals) or too restrictive (rejecting valid ones), motivating expert review or partial specification in high-stakes settings.

Graph construction can be costly and preprocessing-dependent. We treat it as an offline, reusable step (optionally expert-provided), so reported runtime covers only online search; end-to-end time is higher when graph learning is included. In practice, the graph can be refreshed periodically (e.g., after drift) while keeping online explanations lightweight.

Plausibility consistency is implemented via a linear parent-reconstruction penalty: efficient and smooth, but potentially insufficient for nonlinear or context-specific constraints. When conditional probability tables are available, the probabilistic mode better supports discretized/categorical variables and nonlinear conditional relations, but adds modeling choices (discretization, smoothing, CPD sparsity) that affect calibration; a systematic comparison is beyond scope.

We use equal weights for scaled loss terms to avoid per-dataset tuning and preserve comparability across 140 datasets; however, trade-offs among proximity, sparsity, diversity, and plausibility are application-dependent, and deployments may re-weight terms, enforce selected constraints as hard rules, or use domain-specific distance metrics. 

\section{Summary}
\label{sec:summary}

This paper targets a practical bottleneck in counterfactual (CF) explanation: many existing methods treat features as largely independent, which often produces CFs that violate real-world dependencies and therefore appear unrealistic, impractical, or misaligned with domain requirements (e.g., industry standards, audience preferences, or brand guidelines). Such CFs are difficult to act upon, limiting the usefulness of CF-based XAI in operational settings.

We propose DANCE, a dependency-aware CF generation method that explicitly models relationships between features and enforces them during CF search. DANCE represents dependencies as linear or nonlinear constraints whose structure and strength can be learned from data or specified (partially or fully) by domain experts. By integrating these constraints into the generation process, DANCE promotes CFs that better reflect feasible interventions and domain-consistent changes.

Methodologically, DANCE optimizes a composite objective designed to balance key CF qualities---including plausibility, sparsity, and diversity---and employs the Bayesian optimization technique to efficiently discover high-quality CF sets. We evaluate DANCE in two complementary ways: (i) through a real business use case in cybersecurity for online marketing, developed in collaboration with Freshmail within the Sendguard project and deployed in security and marketing operations, and (ii) through a large-scale benchmark on 140 OpenML datasets to assess generality beyond the motivating application. Across these settings, the results demonstrate that making feature dependencies explicit is an effective step toward CF explanations that are more actionable and better aligned with real-world constraints in domains where feasible interventions matter, including healthcare and industrial maintenance~\cite{mtm2025betul,boobek2025tsproto}.

% \section*{Acknowledgments}

%  During the preparation of this work the authors used Microsoft Copilot in order to improve language, fixt spelling mistakes and improve readibility of th emanuscript. After using this tool/service, the authors reviewed and edited the content as needed and takes full responsibility for the content of the published article.

\bibliographystyle{plain}
\bibliography{ccf}

\end{document}